# A Probabilistic Model of Action for Least-Commitment Planning with Information Gathering

**Denise Draper    Steve Hanks    Daniel Weld**
Department of Computer Science and Engineering, FR-35
University of Washington
Seattle, WA 98195
{*ddraper, hanks, weld*}@cs.washington.edu

## Abstract

AI planning algorithms have addressed the problem of generating sequences of operators that achieve some input goal, usually assuming that the planning agent has perfect control over and information about the world. Relaxing these assumptions requires an extension to the action representation that allows reasoning both about the changes an action makes and the information it provides. This paper presents an action representation that extends the deterministic STRIPS model, allowing actions to have both causal and informational effects, both of which can be context dependent and noisy. We also demonstrate how a standard least-commitment planning algorithm can be extended to include informational actions and contingent execution.

## 1 Introduction

The ability to reason with incomplete information, to gather needed information, and to exploit that information in a plan is essential to building agents that can perform competently in realistic domains. Research in AI planning has yielded algorithms for plan generation, but mainly under assumptions that the agent has perfect information about and control over the world.

The decision sciences have developed techniques for representing sources of uncertainty: incomplete information can be viewed as a probability distribution over world states, and conditional probabilities can represent the changes effected by executing an action as well as information gathered during action execution. This formalism provides us with methods for *evaluating* plans, but does not help us to *generate* them.

This paper integrates the two lines of work: we present a representation for actions, plans, and information based on a standard probabilistic interpretation of uncertainty, but one that can also can be manipulated by a subgoaling plan-generation algorithm. The framework allows the representation of information-producing actions (also known as "tests" or "diagnostics"). A standard least-commitment AI planning algorithm is extended to use this probabilistic representation, and further to support contingency plans—plans in which the execution of steps can depend on information provided by previous diagnostic actions. In this paper we will concentrate on the representation for actions and plans, referring the reader to [Draper *et al.*, 1994] for a more detailed description of the planning algorithm.

### 1.1 Example

We begin by posing a simple example that demonstrates the need for reasoning about information, planning to gather information, and acting based on that information.

A robot is given the task of processing a widget. Its goal is to have the widget *painted* (PA) and *processed* (PR) and finally notifying its supervisor that it is done (NO). Processing the widget consists of identifying it as either *flawed* (FL) or *not flawed* ($\overline{\text{FL}}$), then rejecting or shipping the widget (reject or ship), respectively. The robot also has an operator paint that usually makes PA true.

Although the robot cannot immediately tell whether or not the widget is flawed, it does have an operator inspect that tells it whether the widget is *blemished* (BL). The sensor usually reports **bad** if the widget is blemished, and **ok** if not. Initially any widget that is flawed is also blemished. But two things complicate the sensing process:

- Painting the widget removes a blemish but not a flaw, so executing inspect after the widget has been painted conveys no information about whether or not it is flawed.

- The sensor is sometimes wrong: if the widget is blemished then 90% of the time the sensor will report **bad**, but 10% of the time it will erroneously report **ok**. If the widget is not blemished, however, the sensor will always report **ok**.

Initially our robot believes there is a 0.3 chance that the widget is both flawed and blemished (that FL and BL are both true) and a 0.7 chance that it is neither flawed nor blemished.

A classical planner cannot represent this problem, lacking the ability to represent the relative likelihoods of the two possible initial states, the fact that the paint operator can sometimes fail, that the inspect operator provides information about the widget rather than changing its state, and that this information can sometimes be incorrect. The probabilistic planner BURIDAN [Kushmerick et al., 1994a, Kushmerick et al., 1994b] (which cannot create contingent plans), can build a plan with a success probability of at best 0.7: it assumes the widget will not be flawed, paints it, ships it, and notifies its supervisor.

An information-gathering planner can generate a plan that works with probability .97: it first senses the widget, then paints it. Then if the sensor reported ok, it ships the widget, otherwise it rejects it, finally notifying the supervisor. This plan fails only when the widget was initially flawed but the sensor erroneously reports ok, which occurs with probability $(0.3)(0.1) = 0.03$.[1]

The C-BURIDAN planner generates this plan. We will describe the planner by developing this example, first presenting the action and plan representation, then describing a least-commitment planning algorithm that generates probably successful contingent plans.

## 2 States, Actions, and Plans

Here we present the formal definition of a state, an action, a plan, a planning problem, and what it means for a plan to solve a planning problem.

**Propositions and states.** We begin by defining a set of *domain propositions*, each of which describes a particular aspect of the world. Domain propositions for our example are:

FL—widget is flawed      BL—widget is blemished
PR—widget is processed   PA—widget is painted
NO—the supervisor is notified of success

A domain proposition means that aspect of the world is true and a negated domain proposition, e.g. $\overline{FL}$, means that aspect of the world is false. We use the term *literal* to refer to a domain proposition or its negation.

A *state* is a complete description of the agent's model of the world at a particular point in time. Formally we define a state to be a set of literals in which every domain proposition appears exactly once, either negated

or not.[2]

In our example we know that initially the widget has not been processed or painted, and that there is as yet no error. But there is some chance it is both flawed and blemished and some chance it is neither. Thus there are two possible initial states: $s_1 = \{FL, BL, \overline{PR}, \overline{PA}, \overline{NO}\}$ and $s_2 = \{\overline{FL}, \overline{BL}, \overline{PR}, \overline{PA}, \overline{NO}\}$.

We will use $\tilde{s}$ to refer to a random variable over states, and $\tilde{s}_I$ the particular distribution over initial states. This random variable is defined as follows for our example: $(P[\tilde{s}_I = s_1] = 0.3)$, $(P[\tilde{s}_I = s_2] = 0.7)$.

**Expressions.** An *expression* refers to a conjunction of domain literals, which we represent by a set consisting of those literals. The problem's goal is to have the part painted and processed, and supervisor notified—the corresponding expression is $\mathcal{G} = \{PA, PR, NO\}$. The probability that an expression is true in a state is simply:

$$P[\mathcal{E}\,|\,s] \equiv \begin{cases} 1 & \text{if } \mathcal{E} \subseteq s \\ 0 & \text{otherwise} \end{cases} \quad (1)$$

(The probability is 1 if the literals in the expression are all present in the state, 0 otherwise.)

### 2.1 Actions and sequences

An action describes the effects a plan operator has on the world when it is executed. Unlike action representations in classical planners—in which an action's effects are unconditionally realized if the action's precondition is true—the effects of our actions can depend both on the state in which the step is executed as well as random factors (not modeled in the state).

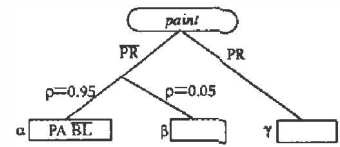

Figure 1: A simple action.

Figure 1 shows a diagram of the paint action: if the widget has already been processed, paint has no effect; otherwise with probability 0.95 the widget will become painted and any blemishes removed and with probability 0.05 the action will not change the state of the world at all.[3]

We describe an action formally as a set of *consequences* $\mathcal{C}_i$. Each consequence $\mathcal{C}_i$ is a tuple of the form $\langle \mathcal{T}_i, \rho_i, \mathcal{E}_i, o_i \rangle$, where $\mathcal{T}_i$ is a set of domain propositions

---

[1] Actually a planner can generate an even better plan by sensing the part repeatedly or by executing paint multiple times.

[2] We define states explicitly in terms of fully specified sets for the sake of formal exposition only—an implementation is not required to represent states this way. In fact, our planning algorithm has no explicit representation of state, instead it reasons directly about the component propositions.

[3] The leaves of the tree indicate *changes* to a state (like STRIPS adds and deletes.)

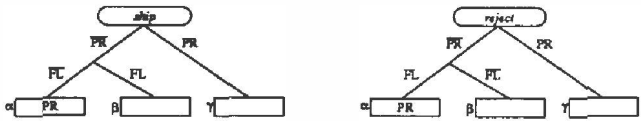

Figure 2: The ship and reject actions

known as the consequence's *trigger*, $\rho_i$ is a probability, $\mathcal{E}_i$ is a set of *effects* associated with the consequence, and $o_i$ is an *observation label* which will be explained below. The idea is that exactly one of an action's consequences is actually realized when the action is executed, and the effects of that consequence determine how the action changes the world. The representation for the paint action pictured in Figure 1 is

paint= { ({PR}, 1.0, {}),
({$\overline{\text{PR}}$}, 0.95, {PA, $\overline{\text{BL}}$}),
({$\overline{\text{PR}}$}, 0.05, {}) }

The *effects* set is a set of literals which describe changes the action makes to the world. We define this formally by the function RES: if $\mathcal{E}$ is an effect set and $s$ is a state,

$$\text{RES}(\mathcal{E}, s) \equiv (s \setminus \{p \mid \overline{p} \in \mathcal{E}\}) \cup \mathcal{E}. \quad (2)$$

It is important to note that these effects describe the *change* an action makes to a state. The paint action cannot possibly make PA false, for example—if its $\beta$ or $\gamma$ consequence occurs, it will not change the world at all. Whether a particular proposition is true after an action is executed can in general be computed only by examining both the action's effects and what state the proposition was in before the action was executed.

An action defines a probabilistic transition from state to states: if $s$ and $w$ are states and A is an action, then

$$P[w \mid s, A] = \sum_{\mathcal{C}_i \in A: w = \text{RES}(\mathcal{E}_i, s)} P[\mathcal{T}_i \mid s] * \rho_i \quad (3)$$

$\sum_w P[w \mid s, A] = 1$ for all states $s$ and all actions A which follows from the fact that (1) distinct trigger expressions are mutually exclusive, so in any state exactly one trigger expression will have probability 1 and the rest will have probability 0, and (2) all probabilities for each individual trigger expression must sum to 1. In other words, an action's consequences are mutually exclusive and exhaustive.

Figure 2 shows two more actions relevant to the example: ship and reject. Ship successfully processes the widget if it is not flawed and not already processed. Reject processes the widget successfully if it *is* flawed and it has not already been processed.

**Action sequences.** We will often reason about executing a *sequence* of actions—we will use $\langle A_1, A_2, \ldots, A_n \rangle$ to mean executing $A_1$, then $A_2$, and so on, and $\langle\rangle$ indicates the execution of no actions. The probability distribution over states induced by executing a sequence of actions is defined as

$$P[u \mid s, \langle\rangle] = \begin{cases} 1 & \text{if } u = s \\ 0 & \text{otherwise} \end{cases} \quad (4)$$

$$P[u \mid s, \langle A_1, A_2, \ldots, A_n\rangle] = \sum_t P[t \mid s, A_1] \times P[u \mid t, \langle A_2, \ldots, A_n\rangle] \quad (5)$$

### 2.2 Information-producing actions

The definition of actions—and the example actions presented so far—have been described in terms of the *changes* they make to the world when they are executed: paint usually makes PA true and makes BL false, for example. What about actions that are executed for the *information* they provide? How does this action representation handle an operator that *finds out* whether the widget is painted (without actually painting it) or the inspect action that determines whether or not the widget is blemished (without either adding a blemish to or removing a blemish from it)?

We cannot use the action's *effects* to model information gathering: doing so would confuse the difference between the changes the action makes and the information it provides, obscuring the difference between a plan that *makes* P true and a plan that *determines whether* P is true [Etzioni *et al.*, 1992]. Instead we model the information produced by an action as a separate report provided to the agent when the action is executed.

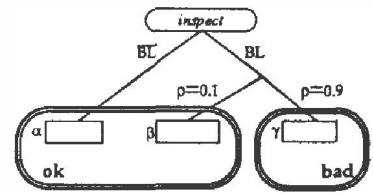

Figure 3: The inspect action provides information but has no material effects.

We divide each action's consequences into a set of *discernible equivalence classes* (or DECs), and assign one report, or *observation label*, to each DEC. When the action is executed, the agent will receive the observation label corresponding to the DEC containing the consequence that was actually realized at execution time. The inspect action (Figure 3) has two DECs: the first consists of the action's $\alpha$ and $\beta$ consequences, and generates the report ok; the second consists of the action's $\gamma$ consequence, and generates the report bad.[4]

The agent gets information from an observation label by making inferences about the world state, reasoning about what consequences of the action could have generated that label. If executing inspect produces the label bad, for example, the agent knows that consequence $\gamma$ occurred, and can be therefore be certain that BL was true when inspect was executed. If it gets the report ok, on the other hand, it knows that either $\alpha$ or $\beta$ occurred, and so it cannot be certain about the state of BL.

---
[4]Note that since all of its effect sets are empty, it will not change the world under any circumstances.

The information generated by executing inspect is summarized by the following conditional probabilities:

$P[\text{bad}|\text{BL}] = 0.9 \quad P[\text{ok}|\text{BL}] = 0.1$
$P[\text{bad}|\overline{\text{BL}}] = 0.0 \quad P[\text{ok}|\overline{\text{BL}}] = 1.0$

which is a standard probabilistic representation of a noisy evidence source (see, e.g., [Pearl, 1988, Chapter 2]).

The updated degree of *belief* in a proposition, conditioned on receiving an observation, is computed as a Bayesian update: suppose inspect is executed in the initial state (where PR is known to be false and $P[\text{BL}] = 0.3$), and the report ok is received:

$$P[\text{BL}|\text{ok}] = \frac{P[\text{ok}|\text{BL}]P[\text{BL}]}{P[\text{ok}|\text{BL}]P[\text{BL}] + P[\text{ok}|\overline{\text{BL}}]P[\overline{\text{BL}}]}$$
$$= \frac{(0.1)(0.3)}{(0.1)(0.3) + (1.0)(0.7)}$$
$$= 0.041$$

We likewise can compute $P[\text{BL}|\text{bad}]$:

$$P[\text{BL}|\text{bad}] = \frac{P[\text{bad}|\text{BL}]P[\text{BL}]}{P[\text{bad}|\text{BL}]P[\text{BL}] + P[\text{bad}|\overline{\text{BL}}]P[\overline{\text{BL}}]}$$
$$= \frac{(0.9)(0.3)}{(0.9)(0.3) + (0.0)(0.7)}$$
$$= 1.0$$

The inspect action can also provide *indirect* evidence—information about propositions other than BL. Since BL and FL are initially perfectly correlated in the example, we have $P[\text{FL}|\text{BL}] = 1$ and therefore can conclude $P[\text{FL}|\text{bad}] = 1$ and $P[\text{FL}|\text{ok}] = 0.041$ as well. Executing paint destroys this correlation, however, so executing inspect after paint would not provide any additional information about FL (but it still would about BL). Thus the information content of an action cannot be fully characterized by examining the action alone—it depends on what probabilistic relationships hold in the plan at the time the action is executed.

Executing an action that has exactly one DEC provides no additional information about the world: the agent knows that one of the consequences occurred, but does not know which one. (We omit the (single) observation label from the action's pictorial representation in such cases, e.g. paint, ship and reject.)

We call an action *information producing* if it has more than one DEC, and *causal* if at least one of its effect sets is non-empty. Information-producing actions correspond to the notions of a *test* or *diagnostic*. Our example actions are either causal but not information producing (e.g. paint) or information producing but not causal (inspect), but our representation allows causal and informational effects to be mixed. This functionality is crucial (and absent from many AI representations of actions with informational effects) because a planner needs to be able to reason about both the benefits and the *costs* of gathering information about the world. In our representation the benefit of sensing is ascertained from the information it produces (its DECs) and the cost of sensing depends on the action's triggers (what must be done to make the sensor operational) and its causal effects (what side effects are generated when the sensing action is executed and how they affect the rest of the plan).

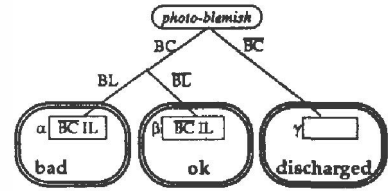

Figure 4: A sensory action with material effects.

Figure 4 shows an example of an action with mixed causal and informational effects. Photo-blemish also detects blemishes, but does so by taking a flash picture. In order for it to take the picture it has to have a charged battery (BC). If the battery is charged the action provides perfect information about the state of BL, and as side effects it illuminates the room (IL) and discharges the battery ($\overline{\text{BC}}$). This action can be used in a plan for a variety of purposes: if the planner can make sure that BC is true, executing the action provides perfect information about BL. The action also provides perfect information about BC and it could be used used to *make* IL true (or BC false). The sensor could be costly if it is difficult to make BC true, or if making IL true has some adverse impact on the rest of the plan.

**Noisy actions.** "Random noise" in the sensing process is handled differently from noise in the effecting process: an effector that fails occasionally (and randomly) is modeled by defining different consequences for the failure and success results. The two consequences have the same trigger, therefore the distinction between the two consequences is made on the basis of the relative probabilities alone. A noisy sensor is modeled by attaching the same observation label to two or more consequences: inspect is a noisy sensor of BL because it can generate the observation label ok both when BL is true and when it is false. An action can be a noisy effector but a perfect sensor, e.g. a pickup action might fail occasionally and probabilistically, but it could provide perfect information about whether or not it succeeded.

**Independence assumptions.** We assume that the conditional probabilities $\rho_i$ of action consequences are independent of one another, both in repeated execu-tion of the same action and execution of different actions. For example, if paint is executed twice (when PR is false) the probability that it will fail to make PA true (at least once) is $1 - .95^2 = .0975$. Likewise, whether or not paint fails does not affect the conditional probability that inspect will fail to recognize a blemished part. Thus we assume that each consequence's conditional probabilities are true with respect to the agent's world model, and we require that each action definition include all dependencies on modeled aspects of the world. For example, suppose that the paint action is more likely to fail if the weather is humid. If humidity is not part of the agent's world model, the contribution of humidity to failure will be encoded in the conditional probability of the $\beta$ consequence (see Figure 1). But if humidity is part of the agent's world model, then the paint action must explicitly represent this dependency (the consequences would be expanded to include the proposition designating humidity in the trigger conditions).

### 2.3 Plan steps and contexts.

A sensory action like inspect becomes useful when the planner can make the execution of other actions contingent on the observation label generated by that sensory action. To represent execution contingency, we define a *plan step*. Where an action defines the effect of an action in abstract, a plan step is a component of a particular plan. A plan step is a pair: $\langle action, context \rangle$. The *action* is as described above, the *context* dictates the circumstances under which the step may be executed.[5] A context is a conjunction of observation labels from previous steps in the plan; a step is executed only if its context matches the observations actually produced during execution.

For example, suppose the agent wishes to execute this sequence of plan steps:

$\langle$ $\langle$inspect, $\{\}\rangle$,
$\langle$ship, $\{$ok$\}\rangle$,
$\langle$reject, $\{$bad$\}\rangle$,
$\langle$notify, $\{\}\rangle$ $\rangle$

Suppose the agent executes the inspect step and receives the report bad. It next considers executing the second step in the sequence, but skips ship since that step's context does not match the report produced by execution of inspect. It does execute the third step, reject, since the step's context *does* match the report produced by inspect. The fourth step, notify, has an empty context so it is executed regardless of whether reject or ship was executed. In summary, the agent must keep track of the *execution context* (the observation labels produced by the steps executed in the plan so far), and execute plan steps only when their context matches the execution context.

### 2.4 Planning problems and solutions

Here we define the probability that a sequence of steps satisfies some goal expression $\mathcal{G}$. The definition is an extension of Equation 5 that takes into account each step's context and its relation to previously executed steps in the sequence. The effect of executing a step given an execution context is either (i) the effect of executing the corresponding action (if the contexts match) otherwise, (ii) no change. Executing an action has two effects: it changes the world state according to its consequences, and also adds an observation label to the execution context.

Let C be the execution context; it is a conjunction of the observation labels that have occurred so far in the plan. The context of step S, context(S), matches C if $C \vdash \text{context}(S)$.

We first define the base case

$$P[u|C, s, \langle\rangle] = \begin{cases} 1 & \text{if } u = s \\ 0 & \text{otherwise} \end{cases} \quad (6)$$

A non-executable step changes neither the distribution over states nor the execution context:

$$P[u|C, s, \langle S_1, S_2, \ldots, S_n\rangle] = P[u|C, s, \langle S_2, \ldots, S_n\rangle] \\ \text{if } C \nvdash \text{context}(S_1) \quad (7)$$

Finally, if step $S_1$ *is* executable in the current context it changes both the probability distribution over states and the execution context, according to the consequences $\mathcal{C}_i = \langle \mathcal{T}_i, \rho_i, \mathcal{E}_i, o_i \rangle$ of action($S_1$):

$$P[u|C, x, \langle S_1, S_2, \ldots, S_n\rangle] = \\ \sum_{\mathcal{C}_i} P[u|(C \wedge o_i), \text{RES}(\mathcal{E}_i, s), \langle S_2, \ldots, S_n\rangle] P[\mathcal{C}_i|s] \\ \text{if } C \vdash \text{context}(S_1) \quad (8)$$

A planning algorithm produces a *solution* to a *planning problem*, both of which we will define now.

A *planning problem* consists of: (1) a probability distribution over initial states $\bar{s}_I$, (2) a *goal expression* $\mathcal{G}$— a set (conjunction) of domain propositions describing the desired final state of the system, (3) a set of *actions* defining the agent's capabilities, and (4) a *probability threshold* $\tau$ specifying a lower bound on the success probability for an acceptable plan.

The planning algorithm produces a sequence of *steps* $\langle S_1, \ldots, S_n \rangle$ as defined above. Such a sequence is a *solution* to the problem if the probability of the goal expression after executing the steps is at least equal to the threshold. The probability of goal satisfaction is defined from Equations (1) and (6)–(8):

$$P[\mathcal{G}|\bar{s}_I, \langle S_1, \ldots, S_n\rangle] = \\ \sum_u P[\mathcal{G}|u]P[u|T, s, \langle S_1, \ldots, S_n\rangle] \sum_s P[\bar{s}_I = s] \quad (9)$$

---

[5]In fact each step also needs a unique *index*, to allow multiple instances of the same action to appear in the plan, in particular allowing the observation labels of repeated actions to be distinguished.

where T is the initial null (always true) execution context. A successful plan is a sequence of steps $\langle S_1, \ldots, S_n \rangle$ that satisfies the inequality $P[\mathcal{G} \mid \bar{s}_I, \langle S_1, \ldots, S_n \rangle] \geq \tau$.

This concludes the formal definition of the problem; next we describe a least-commitment algorithm for solving it.

## 3  Plans and planning

Our planner takes a problem (initial probability distribution, goal expression, threshold, set of actions) as input and produces a solution sequence—a sequence of steps whose probability of achieving the goal exceeds the threshold. Here we describe its data structures and the algorithm it uses to produce a solution. A companion paper [Draper et al., 1994] provides a more detailed description of the algorithm.

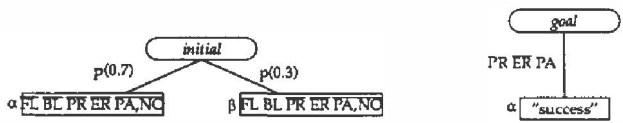

Figure 5: The initial plan

**Initial and goal steps.** The planner initially converts the problem's initial and goal states into two steps, initial and goal. The initial step codes the initial probability distribution, and the goal step has a single consequence with the goal state as its trigger. Figure 5 shows initial and goal actions for the example.

**Plans.** Following BURIDAN [Kushmerick et al., 1993], the planner manipulates a data structure called a *plan*, consisting of a set of *steps*, ordering constraints over the steps, and a set of *causal links*. The initial and goal actions each appear exactly once in every plan, with the initial step ordered before all others and the goal step ordered after all others. A plan with *only* these two steps and this single ordering is called the initial, or null plan, and is the algorithm's starting point.

**Causal links.** Causal links record decisions about the role the plan's steps play in achieving the goal. For example, the planner might create a link from the $\alpha$ consequence of a paint step to the PA trigger proposition of the goal step, indicating that paint is supposed to make PA true for use by goal. We will use the notation $\text{paint}_\alpha \xrightarrow{\text{PA}} \text{goal}$ to refer to this link. The consequence $\text{paint}_\alpha$ is called the link's *producer*, PA is the link's *proposition*, and the step goal is the link's *consumer*.

**Subgoals.** Subgoals are pairs of the form $\langle p, S_j \rangle$ which indicate that the probability of plan success might be increased by increasing the probability of p at $S_j$. Initially, the triggers of goal (*i.e.* the propositions of the goal expression) are subgoals. Thereafter, when a causal link $S_{i\iota} \xrightarrow{p} S_j$ is added to support a subgoal $\langle p, S_j \rangle$, the triggers of consequence $\iota$ of $S_i$ are added to the set of subgoals. In other words, the subgoals are the propositions that participate in chains of causal links ending at the goal. (We will introduce another source of subgoals below.)

**Threats to links.** The process of adding steps and links to the plan can generate conflicts. The presence of a link $S_{i\iota} \xrightarrow{p} S_j$ in a plan actually represents two commitments on the planner's part: (1) to make $S_i$ realize its consequence $\iota$, which will make p true, and (2) to keep p true from $S_i$'s execution until $S_j$'s execution. Therefore a *threat* to the link $S_{i\iota} \xrightarrow{p} S_j$ is any step that (1) possibly occurs between $S_i$ and $S_j$ and (2) has some consequence whose effect set contains $\bar{p}$.

**Planning algorithm.** The planning algorithm can be summarized as follows:

1. Begin with the *null plan*, containing only steps initial and goal, the ordering (initial < goal), and no causal links.

2. Iterate:
   (a) *Assess* the current plan: compute the probability that the current plan achieves the goal. Report success if that probability is at least as great as the threshold.
   (b) Otherwise nondeterministically choose a *refinement* to the current plan (reporting failure if there are no possible refinements), apply the refinement to the current plan, and repeat.

**Assessment.** A plan defines a partial order over its steps, which in turn defines a set of legal execution sequences. One simple assessment algorithm iterates over all step sequences consistent with the plan's ordering constraints, calculating for each totally ordered sequence the set of states that could possibly occur and their associated probabilities (using the definition in Section 2.4), summing the probabilities of all states in which the goal is true. If it finds a sequence with success probability $> \tau$, it returns that sequence, otherwise it returns failure. This simple version of plan assessment is often quite inefficient. [Kushmerick et al., 1993] compares the performance of four different assessment algorithms, including the simple version described here. One of the most interesting assessment algorithms uses the plan's causal links to estimate the success probability without actually enumerating any totally ordered sequences or reasoning explicitly about states.

**Refinement.** A plan refinement adds structure to a plan, trying to increase the probability that the plan will achieve its goal expression. The probability of goal achievement can be increased in one of two ways:

- if $\langle p, S_i \rangle$ is a subgoal, then adding a new link from some (possibly new) plan step that makes p true might increase the probability that p is true at $S_i$, and therefore might increase the success probability,

- if a causal link is currently part of the plan but some other step in the plan threatens the link, then eliminating the threat might increase the probability of the link's consumer proposition, and therefore might increase the success probability.

C-BURIDAN inherits all of BURIDAN's refinement methods (discussed in [Kushmerick et al., 1993, Draper et al., 1994]). We demonstrate them using the example, then describe a new method of threat elimination, *branching*, which introduces contingencies into the plan.

**Example.** Recall that the example problem consists of an initial probability distribution over (two) states and the goal expression {PR, PA, NO}, the actions {paint, reject, ship, inspect}, and a success threshold of $\tau = 0.8$. The initial subgoals are the goal propositions: $\{\langle PR, goal\rangle, \langle PA, goal\rangle, \langle NO, goal\rangle\}$.

The planner can build a non-contingent plan in eight refinement cycles starting from the initial plan (Figure 5). First, the paint step is added along with a link $\text{paint}_\alpha \xrightarrow{PA} \text{goal}$, then $\text{paint}_\alpha$'s trigger $\overline{PR}$ is supported using a link $\text{initial}_\alpha \xrightarrow{\overline{PR}} \text{paint}$. Next the planner supports $\langle PR, goal \rangle$ by adding a ship step, linking its $\alpha$ consequence to the goal, resulting in two new subgoals $\langle \overline{FL}, \text{ship}\rangle$ and $\langle \overline{PR}, \text{ship}\rangle$, both of which can be linked to the initial step's $\alpha$ consequence. The threat that ship poses to the link $\text{initial}_\alpha \xrightarrow{\overline{PR}} \text{paint}$ can then be resolved by ordering paint before ship. Finally, the planner adds a notify step and a link $\text{notify}_\beta \xrightarrow{NO} \text{goal}$, and supports $\text{notify}_\beta$'s trigger PR with the link $\text{ship}_\alpha \xrightarrow{PR} \text{notify}$.

This plan—the best plan a non-contingent planner could produce—will work just in case the widget is initially not flawed and the paint step works, which translates into a success probability of $(0.7)(0.95) = 0.665$. The success probability can be increased somewhat by adding additional paint steps to raise the probability that PA will be true, but without introducing information-producing actions and contingent execution, no planner can do better than 0.7.

At this point a reasonable refinement would be to provide additional support for the subgoal $\langle PR, goal \rangle$ by adding a reject step and linking it to the goal. However this strategy introduces a pair of threats—reject makes PR true, threatening the link from initial to ship, and likewise ship makes PR true, threatening a link from initial to reject—which cannot be solved by adding additional ordering constraints. We need a way to indicate that only *one* of ship or reject should be executed.[6]

---

[6] We have simplified this example so that both ship and reject can actually be executed, but if one succeeds the other will be a no-op. In [Draper et al., 1993] we make ship and reject incompatible by making it an error to execute ship or reject when PR is true.

**Branching.** Branching adds contexts to two plan steps that ensure that the two steps will never both be executable, and therefore that a threat between them will never actually materialize. There are three parts to resolving a threat by branching: (1) choose an information-producing step from the plan (or add a new one) and two disjoint subsets of its observation labels, (2) constrain the execution context of one of the threatening steps to occur only when a label from the first subset is generated and constrain the context of the other threatening step to occur only when a label from the second subset is generated, and (3) generate subgoals for all the triggers of the branching step.

In the example, the planner chooses the information-producing step inspect (adding it to the plan) and restricts the execution context of ship to be **ok** and the execution context of reject to be **bad**. The triggers of inspect can be supported by links from the initial step. Now there is one more threat— the execution of inspect depends on the state of BL, and the execution of paint changes the state of BL—which is easily resolved by ordering inspect before paint. The final plan, Figure 6, has success probability .9215 (it will fail only if the paint step fails or if the widget was blemished and the inspect step incorrectly reports ok) so the planner terminates successfully. The resulting plan is: first inspect the widget, then paint it. If the inspection generated a report of ok then ship the widget, otherwise reject it. Finally, notify the supervisor.

## 4  Conclusion and future work

C-BURIDAN is an implemented algorithm for plan generation that models noisy sensors and effectors according to a standard probabilistic interpretation, but also allows the actions to be manipulated by a symbolic least-commitment planning algorithm. The plan-refinement phase operates on the symbolic part of the action representation, linking (symbolic) action effects to (symbolic) subgoals. The plan-assessment phase treats the actions as probabilistic state transitions, computing a success probability.

The action representation properly distinguishes between an action's causal and informational effects, allowing the planner to discriminate between plans that *make* a proposition true from those that *determine whether* it is true [Etzioni et al., 1992]. The representation makes no arbitrary distinction between sensing actions and effecting actions, however: an action's effects can be both causal and information, and can be noisy in the changes it makes, the information it provides, both, or neither. The representation also allows indirect evidence from sensors to be considered. The plan representation exploits the informational effects of actions by causing execution of steps to be contingent on the observations produced by the execution of previous (information-producing) steps.

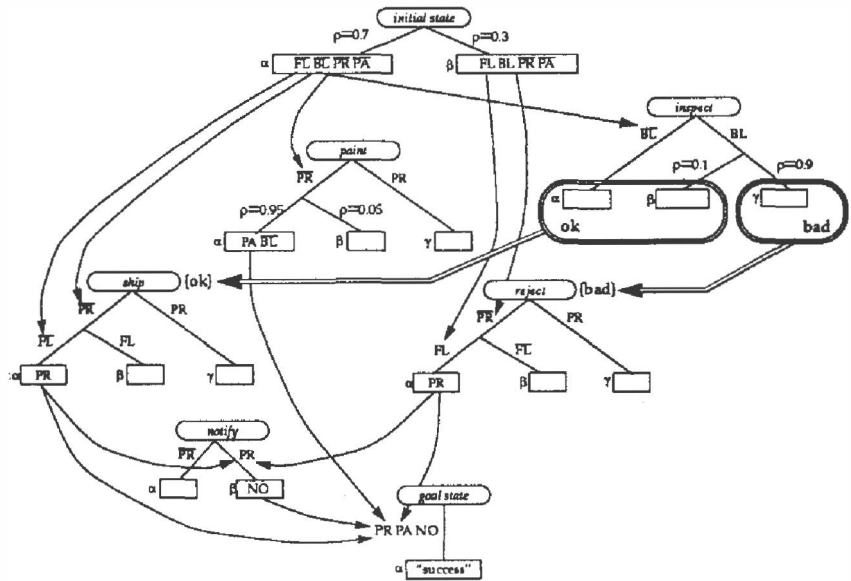

Figure 6: A successful plan $p = .9215$

**Related work.** Related work can be found in the literature on decision making under uncertainty, which deals with evaluating contingency plans with information-gathering actions [Winkler, 1972], [Matheson, 1990]. Also relevant are other symbolic methods for plan-generation under uncertainty [Kushmerick et al., 1993], [Mansell, 1993], [Goldman and Boddy, 1994], and deterministic conditional planners [Peot and Smith, 1992], [Pryor and Collins, 1993]. The longer paper discusses this work in more detail.

Recent work in planning under uncertainty, e.g. [Koenig, 1992] and [Dean et al., 1993], adopts a model based on fully observable Markov processes, which amounts to assuming that the planner is automatically provided with perfect information about the world state every time it executes an action. This assumption is directly opposed to our approach to the problem, in which information about the world is provided only when the agent acts to obtain it, and is potentially incorrect.

Our model of action and information is equivalent in expressive power to a partially observable Markov decision process (POMDP) [Monahan, 1982]. The problem we are solving is different from the one commonly addressed in that literature, however. The POMDP problem is generally posed as finding a *policy* that maximizes some *value function* over some prespecified *horizon*. The horizon is the number of times the policy is to be executed, and may be infinite.

A policy is roughly analogous to our definition of a plan: both tell the agent what to do next based on its prior information about the world and what observations it has received from executing prior actions. It is also straightforward to build a value function that rewards the agent just in case it satisfies a goal expression.

Our planning problem admits no clear notion of a prespecified horizon, however: the agent executes the plan to completion, hoping to satisfy the goal. A horizon is analogous to the number of steps in a plan our algorithm generates, but it is not part of the input problem specification. Further, we do not insist on a policy (plan) that *maximizes* the probability of goal satisfaction, instead accepting any plan that is *sufficiently likely* to satisfy the goal. (Indeed in many cases a finite-length probability-maximizing plan does not exist: if an action fails probabilistically, one can always increase the probability of success by adding another instance of that action to the plan.)

A restatement of our planning problem in the language of POMDP would be "find a policy (with any horizon) that achieves an expected value of at least $v$" where $v$ is some value threshold. We know of no algorithms in the POMDP literature that address this problem.

**Future work.** Future work will be directed in two areas: extending the expressive power of the action representation, and exploring methods for effectively generating contingent plans. The main limitation of the representation language is the absence of any notion of plan *cost*. C-BURIDAN gauges plan success by the probability of satisfying the goal, but [Haddawy and Hanks, 1993] demonstrate the limitations of this model. In order to reason realistically about the cost and value of information, the action representation must be able to handle metric resources (like time, fuel, and money).

As a practical matter, C-BURIDAN can solve only very small problems. The search problem in a probabilistic planner is significantly worse than for a classical planner because the former has to consider the possibility of raising a subgoal's probability by linking to it multiple times. Deciding when to branch and what sensing actions to use also causes computational problems. Ongoing research addresses the problem of how to represent and exploit effective heuristic search-control knowledge.

### Acknowledgments

This research was funded in part by NASA GSRP Fellowship NGT-50822, National Science Foundation Grants IRI-9206733 and IRI-8957302, and Office of Naval Research Grant 90-J-1904.

### References

[Dean et al., 1993] Thomas Dean, Leslie Kaelbling, Jak Kirman, and Ann Nicholson. Planning with deadlines in stochastic domains. In *Proc. 11th Nat. Conf. on A.I.*, July 1993.

[Draper et al., 1993] D. Draper, S. Hanks, and D. Weld. Probabilistic planning with information gathering and contingent execution. Technical Report 93-12-04, University of Washington, December 1993.

[Draper et al., 1994] D. Draper, S. Hanks, and D. Weld. Probabilistic planning with information gathering and contingent execution. In *Proc. 2nd Int. Conf. on A.I. Planning Systems*, June 1994.

[Etzioni et al., 1992] O. Etzioni, S. Hanks, D. Weld, D. Draper, N. Lesh, and M. Williamson. An Approach to Planning with Incomplete Information. In *Proc. 3rd Int. Conf. on Principles of Knowledge Representation and Reasoning*, October 1992. Available via FTP from pub/ai/ at ftp.cs.washington.edu.

[Goldman and Boddy, 1994] Robert P. Goldman and Mark S. Boddy. Representing Uncertainty in Simple Planners. In *Proc. 4th Int. Conf. on Principles of Knowledge Representation and Reasoning*, June 1994.

[Haddawy and Hanks, 1993]
Peter Haddawy and Steve Hanks. Utility Models for Goal-Directed Decision-Theoretic Planners. Technical Report 93-06-04, Univ. of Washington, Dept. of Computer Science and Engineering, September 1993. Submitted to *Artificial Intelligence*. Available via FTP from pub/ai/ at ftp.cs.washington.edu.

[Koenig, 1992] S. Koenig. Optimal probabilistic and decision-theoretic planning using markovian decision theory. UCB/CSD 92/685, Berkeley, May 1992.

[Kushmerick et al., 1993] N. Kushmerick, S. Hanks, and D. Weld. An Algorithm for Probabilistic Planning. Technical Report 93-06-03, Univ. of Washington, Dept. of Computer Science and Engineering, 1993. To appear in *Artificial Intelligence*. Available via FTP from pub/ai/ at ftp.cs.washington.edu.

[Kushmerick et al., 1994a] N. Kushmerick, S. Hanks, and D. Weld. An Algorithm for Probabilistic Least-Commitment Planning. In *Proc. 12th Nat. Conf. on A.I.*, 1994.

[Kushmerick et al., 1994b] N. Kushmerick, S. Hanks, and D. Weld. An Algorithm for Probabilistic Planning. *Artificial Intelligence*, 1994. To appear. Available via FTP from pub/ai/ at ftp.cs.washington.edu.

[Mansell, 1993] T. Mansell. A method for planning given uncertain and incomplete information. In *Proc. 9th Conf. on Uncertainty in Artifical Intelligence*, 1993.

[Matheson, 1990] James E. Matheson. Using Influence Diagrams to Value Information and Control. In R. M. Oliver and J. Q. Smith, editors, *Influence Diagrams, Belief Nets and Decision Analysis*, pages 25-48. John Wiley and Sons, New York, 1990.

[Monahan, 1982] G. E. Monahan. A survey of partially observable markov decision processes: Theory, models, and algorithms. *Management Science*, 28(1):1-16, 1982.

[Pearl, 1988] J. Pearl. *Probablistic Reasoning in Intelligent Systems*. Morgan Kaufmann, San Mateo, CA, 1988.

[Peot and Smith, 1992] M. Peot and D. Smith. Conditional Nonlinear Planning. In *Proc. 1st Int. Conf. on A.I. Planning Systems*, pages 189-197, June 1992.

[Pryor and Collins, 1993] L. Pryor and G. Collins. CASSANDRA: Planning for contingencies. Technical Report 41, Northwestern University, The Institute for the Learning Sciences, June 1993.

[Winkler, 1972] Robert L. Winkler. *Introduction to Bayesian Inference and Decision*. Holt, Rinehart, and Winston, 1972.